\documentclass[10pt,twocolumn,letterpaper]{article}

\usepackage{wacv}

\usepackage{booktabs}
\usepackage{amsmath}
\usepackage{graphicx}
\graphicspath{{figs/}}

\definecolor{wacvblue}{rgb}{0.21,0.49,0.74}
\usepackage[pagebackref,breaklinks,colorlinks,allcolors=wacvblue]{hyperref}
\usepackage[capitalize]{cleveref}

\def\wacvPaperID{61}
\def\confName{WACV}
\def\confYear{2027}

\title{LENS: Likelihood Evaluation and Neighbor Smoothing for\\Attribute-Based Person Retrieval on Unseen Cameras}
\author{Nguyen Dang Duy Tan$^{1}$\quad Vo Thanh Nguyen$^{2}$\quad Linh T. T. Tran$^{3}$\quad Huynh Nguyen Phat$^{4}$\\
$^{1}$AI VIET NAM\qquad $^{2}$Ho Chi Minh City University of Technology and Engineering\\
$^{3}$University of Information Technology, VNU-HCM\qquad $^{4}$University of Science, VNU-HCM\\
{\tt\small nddtan2011@gmail.com}}

\begin{document}
\maketitle

\begin{abstract}
Attribute-based person retrieval ranks surveillance images from a binary query of semantic attributes, but mean Average Attribute Degree of Match (mADM) drops on unseen cameras as predicted probabilities shift. Common pipelines rank gallery images by their L1 distance to the query, which penalizes a confident single-attribute mismatch no more than small uncertainties across many attributes, while post-hoc recalibration fitted on source validation data fails to transfer because the per-attribute maps fitted on source data do not match those of new cameras. We propose LENS (Likelihood Evaluation and Neighbor Smoothing), combining fine-tuned DINOv3 and CLIP models with a distilled student. LENS regularizes mutually exclusive attribute groups with group cross-entropy in training, scores gallery images by the log-likelihood of the exact query, and smooths gallery logits over nearest neighbors using DINOv3 appearance features from the same forward pass. On held-out datasets, all three changes raise mADM: the likelihood score improves all six retrieval metrics for the main ensemble and mADM for all 48 development models, and training-free kNN smoothing improves all 46 tested development models. On the hidden test set of the UPAR 2027 benchmark, LENS reaches 68.17 mADM.
\end{abstract}

\begin{figure*}[t]
\centering
\includegraphics[width=0.87\linewidth]{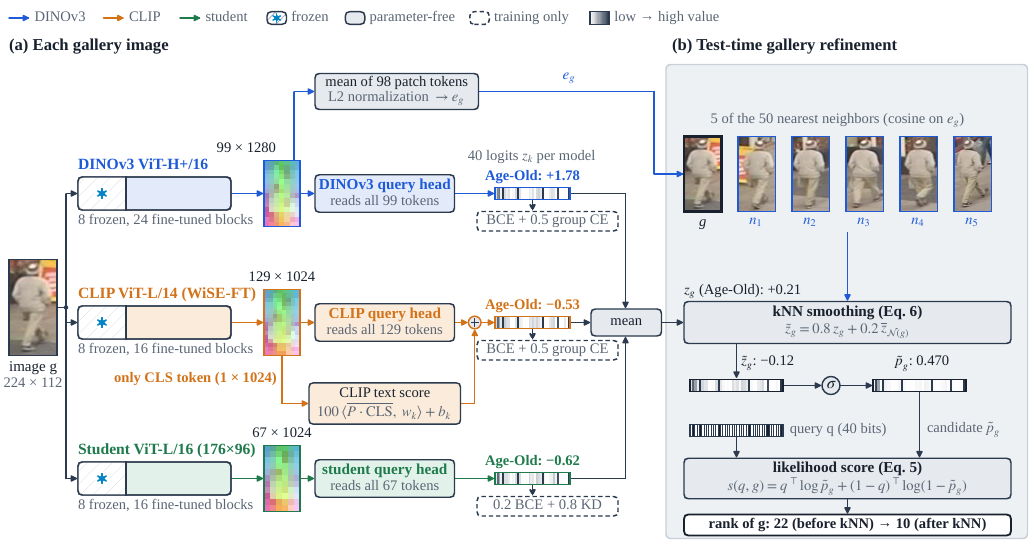}
\caption{Overview of the LENS retrieval pipeline. (a) Three attribute models process one gallery image $g$ from the held-out PETA fold. Each encoder freezes the patch embedding and first eight blocks (snowflake icon) and fine-tunes the remaining blocks with a query head (\cref{fig:qh}). The DINOv3 and CLIP models use BCE and group CE losses, and the student is distilled from two teachers. At inference, their logits are averaged into $z_g$, while the L2-normalized mean of the 98 patch tokens of the DINOv3 model forms $e_g$. (b) Five of the $M=50$ nearest neighbors of $g$ by cosine similarity of $e_g$ (shown here with pretrained DINOv3 features, retrieving the same person in other frames) are averaged in kNN smoothing (\cref{eq:knn}), flipping the Age-Old logit sign to match the label and improving the rank of $g$ under \cref{eq:ll} from 22 to 10.}
\label{fig:arch}
\end{figure*}

\section{Introduction}
\label{sec:intro}

Attribute-based person retrieval finds people in surveillance cameras from a binary query of gender, age, clothing, and accessories without a query image~\cite{upar}. It is usually solved by predicting attribute probabilities for each gallery image and comparing them with the binary query~\cite{upar,specker21}. A central difficulty arises when test images come from cameras never seen in training, because changes in viewpoint, lighting, resolution, and annotation rules between datasets~\cite{upar} can shift the predicted probabilities. In the 2023 and 2024 UPAR challenges~\cite{upar23,upar24}, which trained on PA100k~\cite{pa100k}, PETA~\cite{peta}, and Market-1501~\cite{market,marketattr} and tested on unseen datasets, one team (2023) and ten teams (2024) beat the attribute recognition baseline, but no team beat the retrieval baseline in either edition.

The UPAR 2027 starter kit ranks gallery images by the L1 distance between the binary query and the predicted attribute probabilities~\cite{starterkit}, and earlier work used Euclidean or calibrated, weighted distances~\cite{specker21}. The 2024 report suggested that overfitted models become over- or under-confident~\cite{upar24}. We test both aspects under a leave-one-dataset-out protocol across the three datasets. First, a limitation of L1 distance is that it sums attribute errors linearly, so a confident mismatch on one queried attribute adds no more penalty than small uncertainties spread across several attributes. Ranking instead by the log-likelihood of the exact query~\cite{dap,prp} penalizes near-zero probabilities on queried attributes and raises retrieval mADM on the exact same predicted probabilities without retraining. Second, decomposing negative log-likelihood into calibration and refinement terms~\cite{kull} shows that miscalibration on held-out datasets is real, and comparing per-attribute Platt maps shows that the biases fitted on source validation data are nearly uncorrelated with those of the held-out dataset. Fitting per-attribute Platt scaling~\cite{platt} on source validation splits therefore shifts attribute logits in directions that do not match the new cameras and lowers mADM on both held-out datasets.

Because probability shifts cannot be corrected after training from source validation data in our setting, we propose LENS (\textbf{L}ikelihood \textbf{E}valuation and \textbf{N}eighbor \textbf{S}moothing), shown in \cref{fig:arch}. LENS combines fine-tuned DINOv3~\cite{dinov3} and CLIP~\cite{clip} models, each equipped with a Pre-LN cross-attention query head, together with a distilled student~\cite{hinton,patient}. The CLIP model also adds an attribute text score and interpolates its fine-tuned weights with the pretrained initialization via WiSE-FT~\cite{wiseft}. Because the likelihood score depends on the quality of the predicted probabilities, we regularize mutually exclusive attribute groups with a group cross-entropy loss during training. At inference, we smooth the predicted logits of each gallery image with those of its nearest neighbors in the gallery using DINOv3 appearance features from the same forward pass, an automatic post-processing step that the benchmark rules allow because it uses no knowledge of the hidden test set~\cite{rules}. Our main contributions are:
\begin{itemize}
\item Scoring gallery images by the log-likelihood of the exact query beats L1 distance on all six metrics for the main ensemble and on mADM for every development model.
\item We show that miscalibration on held-out datasets costs mADM, but per-attribute recalibration maps fitted on source data do not match the held-out data, so source-fitted recalibration lowers mADM.
\item Regularizing mutually exclusive attribute groups with group cross-entropy in training raises mADM on both held-out datasets, alone and with kNN smoothing.
\item Smoothing gallery logits over DINOv3 nearest neighbors needs no training and raises mADM for every tested development model on both held-out datasets.
\end{itemize}

\section{Related work}
\label{sec:related}

\paragraph{Attribute-based person retrieval.} Person retrieval by attributes matches surveillance images against a binary attribute query without a query image~\cite{upar,specker21}. Most pipelines predict probabilities for each gallery image and compare them with the query~\cite{upar,specker23}. Specker and Beyerer proposed a calibrated, attribute-weighted, distribution-based distance~\cite{specker21} and the mADM metric~\cite{specker23}, while the UPAR starter kit ranks gallery images by L1 distance~\cite{starterkit}. However, L1 distance sums errors linearly across attributes, penalizing a confident mismatch on one queried attribute no more than small uncertainties across many attributes. By the probability ranking principle~\cite{prp}, ranking by the probability of matching the complete query is optimal when probabilities are accurate; with conditionally independent attributes, this probability is a product of per-attribute probabilities. Direct Attribute Prediction (DAP)~\cite{dap} divides this product by attribute priors, whereas LENS uses the exact-query log-likelihood without this division.

\paragraph{Calibration and attribute dependencies.} Because log-likelihood scoring heavily penalizes near-zero probabilities on queried attributes, its effectiveness hinges on probability reliability under domain shift. While post-hoc methods like Platt~\cite{platt} or temperature scaling~\cite{guo} adjust logits on validation data, calibration maps degrade under distribution shift~\cite{ovadia}. Using the decomposition of proper scoring rules into calibration and refinement~\cite{kull}, we find that held-out datasets are miscalibrated but that per-attribute maps fitted on source data do not match them, so source-fitted calibration hurts retrieval. Rather than adjusting logits post-hoc, LENS enforces probability consistency during training: while the weighted sigmoid cross-entropy of DeepMAR~\cite{deepmar} scores each attribute with an independent sigmoid, we regularize mutually exclusive groups (such as age, hair length, clothing colors~\cite{upar}) with categorical group cross-entropy.

\paragraph{Test-time gallery smoothing.} Single-frame gallery predictions remain vulnerable to viewpoint, blur, or occlusion. In image retrieval and person re-identification, test-time methods refine rankings using gallery structure: $k$-reciprocal re-ranking explores mutual neighbors~\cite{kreciprocal}, query expansion averages top retrieved features~\cite{chum07}, and database-side augmentation combines each database descriptor with those of its neighbors~\cite{turcot,arandjelovic,gordo}. However, these methods require query and gallery images to share a visual feature space, which is absent in attribute-based retrieval where the query is a binary attribute vector. For classification under shift, LAME~\cite{lame} adapts predictions via graph optimization; LENS instead reuses DINOv3 patch tokens from the same forward pass to perform training-free kNN smoothing on attribute logits before scoring.

\section{Method}
\label{sec:method}

\paragraph{Task and evaluation metric.} Each query is a binary vector $q \in \{0, 1\}^K$ over $K$ attributes. Under the UPAR protocol~\cite{upar,specker23}, each distinct ground-truth label vector in the gallery forms a query, and an image $g \in \mathcal{G}$ with binary labels $y_g \in \{0, 1\}^K$ is an exact match when $y_g = q$. Let $\mathrm{dom}(q,g) = \frac{1}{K}\sum_{k=1}^K \mathbb{I}(y_{gk} = q_k)$ be the share of attributes on which $g$ agrees with $q$, normalized relative to the gallery mean $\overline{\mathrm{dom}}_q = \frac{1}{|\mathcal{G}|}\sum_{g' \in \mathcal{G}} \mathrm{dom}(q,g')$ as
\begin{equation}
\mathrm{ndom}(q,g)=\max\Big(0,\tfrac{\mathrm{dom}(q,g)-\overline{\mathrm{dom}}_q}{\max_{g'\in\mathcal{G}}\mathrm{dom}(q,g')-\overline{\mathrm{dom}}_q}\Big).
\label{eq:ndom}
\end{equation}
Let $r_1 < r_2 < \dots < r_R$ be the ranks of the $R$ exact matches in $\mathcal{G}$, and let $g_{(j)}$ be the image at rank $j$. Average Attribute Degree of Match (ADM) is defined as
\begin{equation}
\mathrm{ADM}(q)=\frac1R\sum_{m=1}^{R}\frac{1}{r_m}\sum_{j=1}^{r_m}\mathrm{ndom}\big(q,g_{(j)}\big),
\label{eq:adm}
\end{equation}
and mADM is the mean of $\mathrm{ADM}(q)$ over all queries; we also report mAP, mINP~\cite{reidsurvey}, Rank-1, Rank-5, and Rank-10 (in points). Unlike mAP, which gives zero credit to near matches, mADM takes its score at the exact-match ranks $r_m$ while giving partial credit to near matches ranked ahead of them via $\mathrm{ndom}(q,g_{(j)})$. Consequently, a single confident error on an exact match or an unpenalized mismatch on a distractor pushes exact matches down in rank. LENS addresses this in three connected stages (\cref{fig:arch}): learning non-conflicting attribute logits $z_g$ from localized visual and semantic cues, ranking gallery images by their exact-query log-likelihood rather than L1 distance, and smoothing single-frame logit noise over gallery neighbors.

\begin{figure}[t]
\centering
\includegraphics[width=0.86\linewidth]{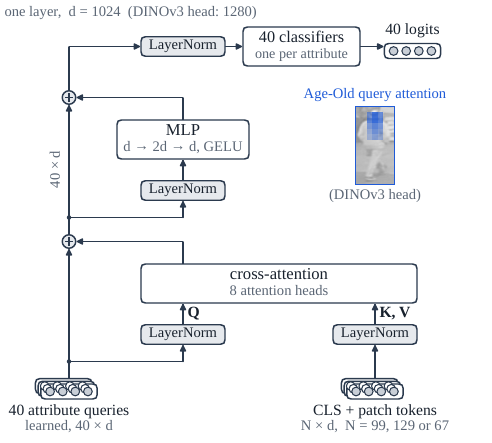}
\caption{Pre-LN cross-attention query head with hidden width $d$ ($d=1280$ for DINOv3; $d=1024$ for CLIP and the student; see \cref{sec:exp}). Inset: Age-Old cross-attention weights over the $14\times 7$ grid of $g$ in \cref{fig:arch}.}
\label{fig:qh}
\end{figure}

\paragraph{Attribute recognition.} Predicting reliable attribute logits on unseen cameras faces two competing demands: small region-level attributes (footwear, headwear, bags) need fine spatial localization, whereas clothing colors and global attributes need semantic invariance to lighting, viewpoint, and camera shifts. We therefore pair two encoders that are pretrained differently: self-supervised DINOv3~\cite{dinov3} learns spatially localized patch tokens that separate fine visual details, whereas CLIP~\cite{clip} learns from image--text pairs and is more robust under distribution shift~\cite{clipdata}, even though it can embed visually distinct images close together~\cite{eyeswide}. LENS therefore pairs a fine-tuned DINOv3 model with a fine-tuned CLIP model (\cref{fig:arch}a). As shown in \cref{fig:arch}a, each encoder freezes its patch embedding and first eight blocks (snowflake icon), fine-tuning the remaining blocks with a one-layer Pre-LN cross-attention query head~\cite{q2l,mldecoder} (\cref{fig:qh}), where $K$ learned attribute queries attend directly to relevant image regions (\cref{fig:qh} inset) before an MLP and linear classifiers output visual logits $z^{\mathrm{vis}} \in \mathbb{R}^K$. To incorporate CLIP's language alignment without overfitting its text encoder, the CLIP model adds a language-anchored text score to its visual query head logit (\cref{fig:arch}a):
\begin{equation}
z_{\mathrm{C},k} = z_{\mathrm{C},k}^{\mathrm{vis}} + 100\,\langle \overline{P \cdot \mathrm{CLS}},\, w_k \rangle + b_k,
\label{eq:clip_text}
\end{equation}
where $P$ is the frozen pretrained projection mapping the image $\mathrm{CLS}$ token to the joint embedding space, the overline denotes $\ell_2$ normalization, $w_k = t_k - t_{\mathrm{neutral}}$ (plus a learnable residual scaled by 0.1) is the prompt-ensembled attribute text embedding minus the neutral prompt ``a person'' (which subtracts shared person-level features and isolates the attribute direction), and $b_k$ is a learnable bias initialized from log-prior odds. Because fine-tuning on source cameras can distort pretrained representations~\cite{lpft}, we interpolate fine-tuned CLIP weights $\theta_{\mathrm{ft}}$ with pretrained weights $\theta_0$ via WiSE-FT~\cite{wiseft}: $\theta = (1-\lambda)\theta_0 + \lambda\theta_{\mathrm{ft}}$ ($\lambda \in [0,1]$).

While pairing DINOv3 and CLIP strengthens the visual and semantic features, standard training with the class-balanced binary cross-entropy of DeepMAR~\cite{deepmar}, $\mathcal{L}_{\mathrm{bce}} = -\sum_{k=1}^K [ \omega_k^{+} y_k \log \sigma(z_k) + \omega_k^{-} (1-y_k) \log(1 - \sigma(z_k)) ]$, treats the $K$ attributes as independent binary tasks on hard labels. This leaves two failure modes for probability-based ranking: independent sigmoids allow mutually exclusive attributes (such as competing age ranges or clothing colors) to fire at once, and hard labels on occluded or low-resolution crops push logits toward overconfident errors. To prevent conflicting predictions within the $S$ mutually exclusive attribute groups $\mathcal{A} = \{A_1, \dots, A_S\}$, we regularize both models in training with a categorical group cross-entropy loss over each group $A \in \mathcal{A}$ that has a single active attribute $y_A \in A$:
\begin{equation}
\mathcal{L}_{\mathrm{gce}}=-\sum_{A \in \mathcal{A}}\mathbb{I}\!\left(\textstyle\sum_{k\in A}y_k=1\right)\log\frac{\exp(z_{y_A})}{\sum_{j\in A}\exp(z_j)},
\label{eq:gce}
\end{equation}
where $\mathbb{I}(\sum_{k\in A}y_k=1)$ skips groups without a unique positive label (such as unlabeled or multi-color clothing), and we optimize $\mathcal{L} = \mathcal{L}_{\mathrm{bce}} + \gamma \mathcal{L}_{\mathrm{gce}}$ ($\gamma > 0$) so attributes within each active group compete. To soften overconfident errors on ambiguous crops and add a third model, we distil a DINOv3 student~\cite{hinton} from DINOv3 and CLIP teachers on mixup~\cite{mixup} and CutMix~\cite{cutmix} images~\cite{patient}, minimizing $\mathcal{L}_{\mathrm{student}} = (1-\beta)\mathcal{L}_{\mathrm{label}} + \beta\mathcal{L}_{\mathrm{KD}}$ against the teacher ensemble probabilities $\bar{p}_t = \sigma(\tfrac{1}{2}(z_{\mathrm{D}} + z_{\mathrm{C}}))$. At inference, the logits of DINOv3, CLIP, and the student are averaged into $z_g = \frac{1}{3}(z_{\mathrm{D}} + z_{\mathrm{C}} + z_{\mathrm{S}}) \in \mathbb{R}^K$ (\cref{fig:arch}a), giving candidate probabilities $p_g = \sigma(z_g)$ without tuned thresholds.

\paragraph{Exact-query likelihood scoring.} Given the predicted probabilities $p_g = \sigma(z_g)$ for each gallery image $g \in \mathcal{G}$, the starter kit ranks the gallery by the L1 distance $d(q, g) = \sum_{k=1}^K |q_k - p_{gk}|$~\cite{starterkit}. Because $d(q,g)$ equals the expected number of mismatched attributes, it sums errors linearly and cannot separate one confident mismatch from many small uncertainties: a distractor matching $K-1$ attributes that is confidently wrong on one queried attribute ($p_{gk} \approx 0$ for $q_k = 1$) adds less than $1$ to $d(q,g)$ and can rank ahead of an exact match with slight uncertainty across several attributes. Because mADM (\cref{eq:adm}) scores the ranking at the positions of exact matches, gallery images should instead be ordered by the probability that all $K$ attributes match $q$ at once. By the probability ranking principle~\cite{prp}, this order is optimal for exact-match relevance when the probabilities are accurate and attributes are conditionally independent given the image. DAP~\cite{dap} divides each posterior by its attribute prior $\pi_k$; in log-space, the prior term depends only on $q$ and cancels across $g \in \mathcal{G}$, whereas shifting logits by source log-priors before $\sigma(\cdot)$ distorts $\sum_{k=1}^K \log(1-p_{gk})$ under prior shift. LENS ranks candidates by the exact-query log-likelihood against the 40-bit query $q$ (\cref{fig:arch}b):
\begin{equation}
s(q,g)=\sum_{k=1}^K\big[q_k\log p_{gk}+(1-q_k)\log(1-p_{gk})\big],
\label{eq:ll}
\end{equation}
which simplifies via $z_{gk} = \log \frac{p_{gk}}{1-p_{gk}}$ to $s(q,g) = q^\top z_g + \sum_{k=1}^K \log(1-p_{gk})$, reducing scoring all queries against the gallery to one matrix product $Q Z^\top$ plus one constant per image.

\begin{table}[t]
\centering
\small
\setlength{\tabcolsep}{1.3pt}
\begin{tabular}{@{}llcccccc@{}}
\toprule
Fold & Score & mADM & mAP & mINP & R-1 & R-5 & R-10 \\
\midrule
PETA & L1 distance~\cite{starterkit} & 62.88 & 24.03 & 18.67 & 21.75 & 39.59 & 47.22 \\
& likelihood (\ref{eq:ll}) & \textbf{69.91} & \textbf{35.51} & \textbf{29.08} & \textbf{31.57} & \textbf{54.68} & \textbf{63.07} \\
& likelihood$^\dagger$ & 69.84 & 35.38 & 28.99 & 31.59 & 54.17 & 63.30 \\
& group-softmax$^\dagger$ & 68.62 & 35.47 & 28.77 & 32.10 & 54.34 & 63.41 \\
\midrule
Market & L1 distance~\cite{starterkit} & 69.29 & 25.08 & 8.97 & 32.53 & 51.46 & 58.00 \\
& likelihood (\ref{eq:ll}) & \textbf{74.90} & \textbf{34.39} & \textbf{16.84} & \textbf{37.97} & \textbf{61.30} & \textbf{70.87} \\
& likelihood$^\dagger$ & 74.59 & 34.17 & 16.48 & 38.28 & 60.77 & 71.13 \\
& group-softmax$^\dagger$ & 73.63 & 34.38 & 16.57 & 39.64 & 61.92 & 71.97 \\
\bottomrule
\end{tabular}
\caption{Scoring rules on the DINOv3 + CLIP ensemble (CLIP with text score and WiSE-FT), in points. Bold marks the better of L1 distance and likelihood. $^\dagger$Same single-seed logits as \cref{tab:gce}.}
\label{tab:score}
\end{table}

\paragraph{kNN smoothing.} Because the likelihood score (\cref{eq:ll}) heavily penalizes confident errors, an exact match $g$ whose logit on one attribute is flipped by single-frame occlusion, blur, or angle suffers a steep rank drop. The test-time gallery refinement stage in \cref{fig:arch}b addresses this vulnerability: surveillance images are not independent because a gallery image $g$ shares visual neighbors of two kinds, namely other frames of the same person and different individuals with matching clothing and appearance. Averaging logits across such visually consistent neighbors suppresses single-frame noise without labels or queries. Rather than running a separate network, we reuse the localized patch tokens of the DINOv3 encoder from the same forward pass (\cref{fig:arch}a), defining appearance feature $e_g$ as the L2-normalized mean of its 98 patch tokens. Let $\mathcal{N}(g)$ be the $M$ nearest neighbors of $g$ in $\mathcal{G}\setminus\{g\}$ by cosine similarity of $e_g$ (\cref{fig:arch}b). We smooth the ensemble logits over $\mathcal{N}(g)$ before scoring:
\begin{equation}
\widetilde{z}_g=(1-\alpha)\,z_g+\frac{\alpha}{M}\sum_{n\in\mathcal{N}(g)}z_n,
\label{eq:knn}
\end{equation}
and substitute $\widetilde{z}_g$ (with candidate probabilities $\widetilde{p}_g = \sigma(\widetilde{z}_g)$) into \cref{eq:ll}.

\section{Experiments}
\label{sec:exp}

\begin{figure*}[t]
\centering
\includegraphics[width=\linewidth]{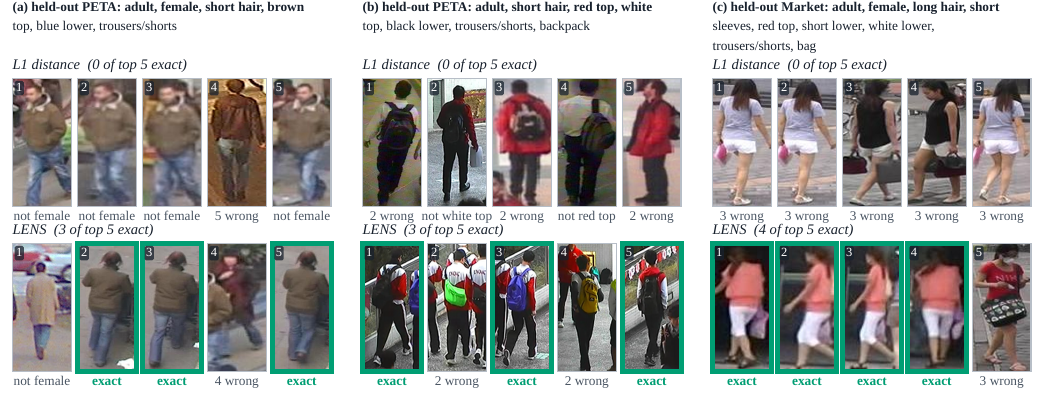}
\caption{Top five gallery images of three held-out queries under the L1 distance and under LENS (likelihood score with kNN smoothing). Both rankings start from the same probabilities of the DINOv3 + CLIP models of the fold (trained with group CE on the two other datasets, CLIP without WiSE-FT), and kNN smoothing uses the fine-tuned DINOv3 features of the fold ($M=50$, $\alpha=0.2$). Green borders mark exact matches, and the labels under the other images give the single mismatched attribute or the number of mismatched attributes. The queries were selected among those with 2 to 8 exact matches for which the L1 distance has no exact match in its top five and LENS has at least three.}
\label{fig:qual}
\end{figure*}

\paragraph{Evaluation protocol.}
\label{sec:protocol}
We use a leave-one-dataset-out protocol across PA100k~\cite{pa100k}, PETA~\cite{peta}, and Market-1501~\cite{market,marketattr}: models train on two source training splits and are evaluated on the held-out validation gallery, where each distinct label vector forms a query. Held-out PETA has 6,963 images and 1,763 queries (89.0k source training images from PA100k + Market-1501); held-out Market-1501 has 16,458 images and 956 queries (87.7k source images from PA100k + PETA). Holding out PA100k (9,986 images, 1,479 queries) leaves only 18.7k combined source training images from PETA + Market-1501 (nearly five times fewer than the other folds), so we train full multi-encoder ensembles on the PETA and Market-1501 folds and include two PA100k-fold models in our 48 development models spanning ViT and ConvNeXt encoders at sizes B, L, and H (41 on PETA, 5 on Market-1501, and 2 on PA100k). Replicate standard deviation across seeds is $0.15$ mADM points (paired bootstrap standard error of a difference $0.24$ to $0.29$); single-seed runs in \cref{tab:calib,tab:gce} match two-seed CLIP averages in \cref{tab:score,tab:enc} within $\le 0.31$ points.

\begin{table}[t]
\centering
\small
\setlength{\tabcolsep}{4pt}
\begin{tabular}{@{}lcc@{}}
\toprule
& PETA & Market \\
\midrule
NLL on the held-out dataset (nat/image) & 6.20 & 5.26 \\
NLL on source datasets (nat/image) & 4.48 & 4.28 \\
Calibration term, held-out dataset & 1.11 & 1.01 \\
Calibration term, source datasets & 0.80 & 1.06 \\
\midrule
mADM, no recalibration & 69.53 & 74.66 \\
mADM, oracle recalibration & 71.95 & 76.26 \\
mADM, source recalibration & 68.51 & 73.65 \\
mADM, logits divided by $T=0.9$ & 69.16 & 74.44 \\
mADM, logits divided by $T=1.1$ & 69.64 & 74.79 \\
mADM, logits divided by $T=1.3$ & 69.49 & 74.82 \\
\bottomrule
\end{tabular}
\caption{Recalibration after training for the DINOv3 + CLIP ensemble (seed 1, CLIP with text score, without WiSE-FT) using $\sigma(a_k z + b_k)$ per attribute. Oracle recalibration fits on held-out dataset labels, and source recalibration fits on source validation images. The calibration term~\cite{kull} is the NLL removed by the per-attribute Platt map fitted on the same images (a lower bound).}
\label{tab:calib}
\end{table}

\paragraph{Implementation details.}
\label{sec:impl}
The benchmark uses $K=40$ binary attributes across PETA, Market-1501, and PA100k, with five mutually exclusive groups: three age ranges, three hair lengths, two lower-body clothing types, 12 upper colors, and 12 lower colors. The encoders are DINOv3 ViT-H+/16~\cite{dinov3} (840M parameters, query head width $d=1280$) and OpenAI CLIP ViT-L/14~\cite{clip} (303M parameters, $d=1024$), both trained at $224\times112$. The student is DINOv3 ViT-L/16 (304M, $d=1024$) at $176\times96$ (for CPU-only inference, DINOv3 ViT-B/16, 86M, $d=768$ at $256\times128$), distilled with $\beta=0.8$ (50\% mixup~\cite{mixup}/CutMix~\cite{cutmix}). Encoders freeze the patch embedding and first eight blocks, fine-tuning remaining blocks on a single GPU (RTX 5090 or RTX PRO 6000) with AdamW ($\beta_1=0.9, \beta_2=0.999$, weight decay $0.05$), batch size 64, and EMA. Base learning rate is $5\times 10^{-5}$ at the top block (0.75 layer decay below) and $5\times 10^{-4}$ in the query head. WiSE-FT on CLIP uses $\lambda=0.7$. Training combines BCE with DeepMAR weights $\omega_k$~\cite{deepmar}, label smoothing $\epsilon = 0.05$~\cite{szegedy}, and group CE weight $\gamma = 0.5$. Teachers train for 3 epochs and the student for 20 epochs. At inference, kNN smoothing uses DINOv3 patch features with $M=50, \alpha=0.20$ (0.1\,s on GPU for 6,963 images), and probabilities are clipped to $[10^{-4}, 1-10^{-4}]$ (logits within $\pm 9.2$) before \cref{eq:ll}, bounding single-attribute penalty at $-\log 10^{-4} \approx 9.2$ nats.

\paragraph{Scoring gallery images.}
\label{sec:score}

We first measure how much the scoring rule changes the ranking when predicted probabilities are fixed. \Cref{tab:score} compares the L1 distance of the starter kit~\cite{starterkit} with the likelihood score (\cref{eq:ll}) on the same DINOv3 + CLIP ensemble. The likelihood score raises mADM by 7.03 points on PETA and 5.61 points on Market-1501, and increases all five other metrics. This result repeats across all 48 development models, beating L1 on every model by 3.7 to 9.1 mADM ($+8.16$ and $+9.10$ on the two held-out PA100k models).

\begin{figure}[t]
\centering
\includegraphics[width=0.98\linewidth]{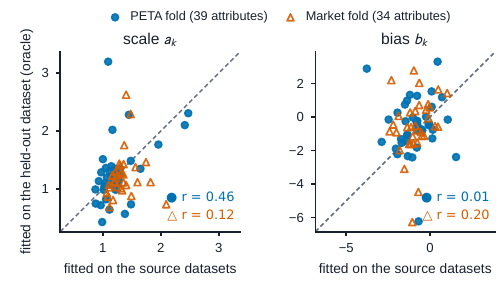}
\vspace{-2pt}
\caption{Per-attribute Platt parameters $\sigma(a_k z + b_k)$ fitted on source vs.\ held-out labels for the ensemble in \cref{tab:calib} ($y=x$ dashed, Pearson $r$, omitting attributes without both classes on the held-out split).}
\label{fig:calib}
\end{figure}

\begin{table}[t]
\centering
\small
\setlength{\tabcolsep}{4pt}
\begin{tabular}{@{}lccc@{}}
\toprule
Model & PETA & Market & Mean \\
\midrule
DINOv3 & 66.36 & 73.87 & 70.12 \\
DINOv3, two fine-tuned copies & 66.69 & 74.00 & 70.34 \\
CLIP & 65.81 & 71.72 & 68.76 \\
CLIP + text score & 67.17 & 72.32 & 69.74 \\
CLIP + text score + WiSE-FT & 67.38 & 71.98 & 69.68 \\
\midrule
DINOv3 + CLIP & 68.92 & 74.46 & 71.69 \\
DINOv3 + CLIP (text score) & 69.49 & 74.79 & 72.14 \\
DINOv3 + CLIP (text score, WiSE-FT) & \textbf{69.91} & \textbf{74.90} & \textbf{72.40} \\
\bottomrule
\end{tabular}
\caption{Held-out mADM (points) of individual models and ensembles under the likelihood score. The second row averages two DINOv3 models fine-tuned independently with the same recipe. The last row is the final setting without group CE, kNN smoothing, and the student.}
\label{tab:enc}
\end{table}

\begin{figure}[t]
\centering
\includegraphics[width=0.98\linewidth]{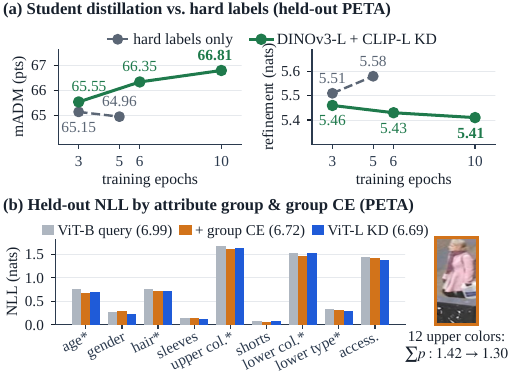}
\caption{Improving predicted probabilities on held-out PETA. (a) Training the DINOv3 ViT-L/16 student ($176\times 96$) on hard labels alone vs.\ distilling from DINOv3-L + CLIP-L teachers across epochs. (b) Held-out NLL per image by attribute group (asterisks mark groups in \cref{eq:gce}), with a PETA crop whose 12-color probability sum drops from $1.42$ to $1.30$.}
\label{fig:enckd}
\end{figure}

\Cref{fig:qual}a shows the cause on a query for an adult woman. Four of the five images that L1 distance ranks first are frames of one man: the model predicts he is not female ($p=0.03$ to $0.04$) while all 39 other attributes match. This single confident mismatch adds less than 1 to the L1 distance, smaller than the sum of slight uncertainties across 40 attributes on exact matches, whereas $-\log p$ adds 3.2 to 3.6 nats. LENS strongly penalizes the near-zero female probability and places three exact matches in the top five. In the other two queries of \cref{fig:qual}, L1 distance also has no exact match in its top five, whereas LENS places three and four. Against other scoring rules, literal log-DAP~\cite{dap} subtracts a query-only prior constant ($\Delta\text{mADM} = 0.00$), and weighting attributes by source AUC, a simplified form of the weighting in~\cite{specker21}, changes mADM by $0.00/+0.02/-0.40$ on PETA/Market-1501/PA100k for earlier single models. On the same logits, replacing sigmoids in each mutually exclusive group by $\log\mathrm{softmax}(z_A)$ (group-softmax in \cref{tab:score}) slightly raises mAP and Rank-$k$ but lowers mADM by 1.22 and 0.96 points, plausibly because it forces one positive even when a group has zero or two active labels.

\paragraph{Post-hoc recalibration on unseen cameras.}
\label{sec:calib}

We test whether models are miscalibrated on new cameras with per-attribute Platt scaling~\cite{platt}, mapping each logit via $\sigma(a_k z + b_k)$ in two ways: oracle recalibration fits $a_k, b_k$ on held-out labels to measure the mADM cost of miscalibration, whereas source recalibration fits $a_k, b_k$ on the validation splits of the two source datasets (using the seed-1 DINOv3 + CLIP ensemble without WiSE-FT, which saved source validation logits).

\Cref{tab:calib} shows that held-out miscalibration is real: oracle recalibration raises mADM by 2.42 points on PETA and 1.60 points on Market-1501. However, source recalibration lowers mADM by about one point on both folds, and lowers mADM on 47 of the 48 development models (except one linear-head model on the smaller PA100k fold).

The cause is not that miscalibration appears only on new cameras: on the source datasets, the calibration term is 72\% of the held-out value in the PETA fold and exceeds it in the Market-1501 fold. Rather, the per-attribute maps differ between the two fits (\cref{fig:calib}): the scale $a_k$ is above 1 for most attributes on both splits, but the per-attribute values correlate weakly (scale $r = 0.46$ and $0.12$, bias $r = 0.01$ and $0.20$ on PETA and Market-1501). Source-fitted maps therefore shift logits in directions that do not match the new cameras. A single shared temperature $T$ changes mADM by at most 0.4 points on PETA and Market-1501 (but by $-0.42$ and $-1.34$ at $T=1.1$ and $1.3$ for one earlier PA100k-fold model). Miscalibration costs mADM, but source validation data cannot guide its correction, so LENS uses uncalibrated logits and improves probabilities via stronger models, group CE, and kNN smoothing.

\begin{table}[t]
\centering
\small
\setlength{\tabcolsep}{2.5pt}
\begin{tabular}{@{}lccc@{}}
\toprule
DINOv3 + CLIP ensemble & PETA & Market & Mean \\
\midrule
Likelihood score & 69.84 & 74.59 & 72.22 \\
+ kNN smoothing & 70.23 & 75.31 & 72.77 \\
+ group CE & 70.89 & 75.28 & 73.08 \\
+ group CE + kNN smoothing (final) & \textbf{71.03} & 75.66 & 73.35 \\
\midrule
\quad kNN with pretrained DINOv3 & 71.00 & 75.52 & 73.26 \\
\quad row 4 + query expansion & 70.99 & \textbf{75.94} & \textbf{73.47} \\
\bottomrule
\end{tabular}
\caption{Held-out mADM (points) of the DINOv3 + CLIP ensemble (seed 1) under the likelihood score. Group CE is applied to both models, and kNN smoothing uses $M=50$, $\alpha=0.2$, and fine-tuned DINOv3 features except in the fifth row. The fourth row is the final setting without the student; the first row differs slightly from the last row of \cref{tab:enc} because this table uses CLIP seed 1 only and a retrained DINOv3 model.}
\label{tab:gce}
\end{table}

\begin{table}[t]
\centering
\small
\setlength{\tabcolsep}{3.2pt}
\begin{tabular}{@{}lcccc@{}}
\toprule
Neighbor feature & $M$, $\alpha$ & PETA & Market & PA100k \\
\midrule
no smoothing (mADM) & & 68.45 & 78.14 & 73.86 \\
\midrule
DINOv3, mean patch token & 3, 0.25 & $-$0.03 & +0.67 & +0.52 \\
DINOv3, mean patch token & 10, 0.25 & +0.56 & +0.86 & +0.67 \\
DINOv3, mean patch token & 20, 0.25 & +0.64 & +0.86 & +0.65 \\
DINOv3, mean patch token & 50, 0.25 & +0.60 & +0.85 & +0.70 \\
DINOv3, mean patch token & 50, 0.2 & +0.70 & +0.69 & +0.66 \\
DINOv3, mean patch token & 20, 0.5 & $-$1.58 & +0.82 & $-$0.29 \\
DINOv3, CLS token & 20, 0.25 & +0.54 & +0.64 & +0.78 \\
CLIP, mean patch token & 20, 0.25 & +0.23 & +1.07 & +0.29 \\
\bottomrule
\end{tabular}
\caption{Change in mADM (points) from \cref{eq:knn} relative to the first row on an earlier DINOv3 + CLIP ensemble of the PETA fold (without text score) across the unseen target gallery (held-out PETA) and the two source-domain validation galleries (Market-1501 and PA100k, excluded from training).}
\label{tab:knn}
\end{table}

\begin{figure}[t]
\centering
\includegraphics[width=0.98\linewidth]{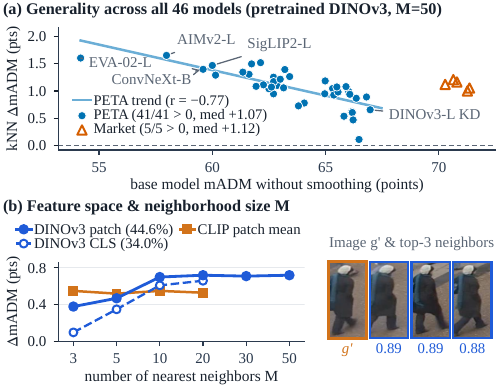}
\caption{Generality and feature analysis of kNN smoothing. (a) $\Delta$mADM across all 46 development models with saved gallery features ($M=50$, $\alpha=0.2$) using pretrained DINOv3 features: all 41 PETA models ($[+0.11, +1.65]$) and all 5 Market-1501 models ($[+1.00, +1.21]$) improve. (b) Mean $\Delta$mADM across the three galleries in \cref{tab:knn} vs.\ $M$ ($\alpha=0.25$), with a PETA crop $g'$ and top-3 neighbors.}
\label{fig:knn}
\end{figure}

\paragraph{The role of two different encoders.}
\label{sec:enc}

Because probabilities cannot be fixed after training, we improve the probabilities themselves (\cref{tab:enc}). On held-out PETA under \cref{eq:ll}, a DINOv3-pretrained ConvNeXt-B~\cite{convnext} with a query head (\cref{fig:knn}a) reaches 59.59 mADM (62.26 for ConvNeXt-L), and single-encoder query-head models with AIMv2-L~\cite{aimv2}, SigLIP2-L~\cite{siglip2}, or MetaCLIP2-H~\cite{metaclip2}, trained at lower input resolutions, reach 57.98, 60.00, and 63.15, below DINOv3 ViT-H+/16 (66.36). Averaging two fine-tuned copies of DINOv3 raises the two-fold mean mADM by only 0.22 points over a single copy, whereas pairing DINOv3 with CLIP raises the mean by 1.57 points even though CLIP alone is weaker than DINOv3, showing that pairing two differently pretrained encoders helps more than averaging two copies of one encoder. Adding the CLIP text score and WiSE-FT raises the ensemble by another 0.71 points in the mean, mostly on PETA.

To isolate distillation across training length, \cref{fig:enckd}a trains the DINOv3 ViT-L/16 student on the PETA fold using DINOv3 ViT-L/16 and CLIP ViT-L/14 teachers. At 3 epochs, distillation raises held-out PETA mADM by 0.40 points over hard labels alone (65.55 vs.\ 65.15). Extending hard-label training from 3 to 5 epochs does not improve mADM (64.96) and worsens refinement from 5.51 to 5.58 nats (even as total NLL drops from 6.91 to 6.77 nats), whereas distillation keeps improving the student through 10 epochs to 66.81 mADM. Adding the 10-epoch ViT-L/16 student to the DINOv3 + CLIP ensemble raises PETA mADM from 69.91 to 70.24.

\paragraph{Group CE and kNN smoothing.}
\label{sec:gce}

Because the likelihood score depends on probability quality, we regularize mutually exclusive groups in training with group CE (\cref{fig:enckd}b). \Cref{tab:gce} evaluates group CE and kNN smoothing on both held-out datasets: group CE raises mADM by 1.05 points on PETA and 0.69 points on Market-1501, and also improves the DINOv3 model alone from 65.95 to 67.25 on PETA and from 73.38 to 74.41 on Market-1501. Meanwhile, kNN smoothing raises mADM by 0.39 and 0.72 points alone, and adds 0.14 (within noise) and 0.38 points after group CE. Pretrained DINOv3 features without fine-tuning give nearly the same gain on PETA and about two thirds of the gain on Market-1501, showing that the neighbor features need no task fine-tuning. Query expansion~\cite{chum07} raises mADM on Market-1501 (75.94) but not on PETA (70.99) and lowers Rank-5 and Rank-10 on both folds, so we omit it.

\begin{table}[t]
\centering
\small
\setlength{\tabcolsep}{4pt}
\begin{tabular}{@{}p{0.8\linewidth}c@{}}
\toprule
Final model & mADM \\
\midrule
DINOv3 ViT-B/16 student & 63.65 \\
DINOv3 + CLIP + DINOv3 ViT-L/16 student & 67.22 \\
\quad + text score, WiSE-FT, kNN smoothing & 67.83 \\
\quad + group CE for DINOv3 and CLIP & \textbf{68.17} \\
\bottomrule
\end{tabular}
\vspace{-3pt}
\caption{mADM (points) on the hidden test set of the UPAR 2027 benchmark under the likelihood score. All models are trained on the training and validation splits of the three datasets.}
\label{tab:test}
\end{table}

\paragraph{Sensitivity of kNN smoothing.}
\label{sec:knn}

\Cref{tab:knn} varies $M$, $\alpha$, and the neighbor feature. On the two source-domain validation galleries, $M \in [10, 50]$ and $\alpha \in [0.2, 0.25]$ form a broad plateau, and every setting on it transfers positively to held-out PETA ($+0.56$ to $+0.70$ in \cref{tab:knn}, $+0.52$ with the student at $M=50, \alpha=0.2$, and $+0.73$ to $+0.83$ on held-out Market-1501 for the Market-fold ensemble; only $\alpha=0.5$ over-smooths PETA by $-1.58$). We fix $M=50$ and $\alpha=0.2$ from this source-validation plateau.

$M=3$ already helps on Market-1501 ($+0.67$) and PA100k ($+0.52$) but not on PETA ($-0.03$). Under DINOv3 appearance features, the top-1 neighbor shares all 40 ground-truth labels with the image for 76.7\% of Market-1501, 50.6\% of PA100k, and 44.6\% of PETA images (\cref{fig:arch}b, \cref{fig:knn}b), as 85\% of Market-1501 queries have more than one exact match in the gallery whereas 61\% of PETA queries have a single one. On Market-1501, the few nearest neighbors are mostly frames of the same person; on PETA, which has fewer repeated identities, $M=50$ likely averages over different pedestrians with similar clothing, whereas neighbors found by predicted probability cosine similarity ($M=50, \alpha=0.2$) lower PETA mADM in \cref{tab:score} by 0.13 points ($+0.17$ on Market-1501).

With frozen pretrained DINOv3 features, kNN smoothing needs no training and raises mADM for all 46 development models with saved gallery features (\cref{fig:knn}a), by a median of $+1.07$ points on PETA and $+1.12$ on Market-1501. The gain shrinks as the base model gets stronger ($r = -0.77$ on PETA), consistent with \cref{tab:gce}.

\paragraph{Evaluation on the UPAR 2027 hidden test set.}
\label{sec:test}

For the hidden test set, teacher and student models train on the combined training and validation splits of PA100k, PETA, and Market-1501 with the fold hyperparameters (\cref{tab:test}). For CPU-only inference, a standalone DINOv3 ViT-B/16 model with group CE reaches 62.47 mADM, rising to 63.15 with 10-epoch distillation and 63.65 with 20-epoch mixup/CutMix distillation. Combining DINOv3 ViT-H+/16, CLIP ViT-L/14, and the DINOv3 ViT-L/16 student (\cref{fig:arch}) reaches 67.22 mADM; reusing the DINOv3 and student weights while adding the CLIP text score, WiSE-FT, and kNN smoothing reaches 67.83, and retraining both teachers with group CE ($+0.34$) reaches \textbf{68.17 mADM}.

\section{Limitations and future work}
\label{sec:limits}

Our experiments cover PA100k, PETA, Market-1501, and the UPAR 2027 hidden test set with 40 binary attributes, with one or two seeds per fold model. Future work will test open-vocabulary queries and other benchmarks, unsupervised test-time calibration beyond Platt and temperature scaling, and neighbor weights that adapt to how often a person reappears in the gallery. Given privacy risks in surveillance retrieval, we use only public benchmarks.

\section{Conclusion}
\label{sec:conclusion}

We studied attribute-based person retrieval on unseen cameras across PA100k, PETA, Market-1501, and UPAR 2027. Ranking gallery images by the exact-query log-likelihood outperforms the L1 distance on the same predicted probabilities because it penalizes confident single-attribute mismatches. Held-out miscalibration is real, but its per-attribute direction cannot be estimated from source data, so LENS instead improves the probabilities via DINOv3 and CLIP teachers, student distillation, group CE, and kNN smoothing, reaching 68.17 mADM on the UPAR 2027 hidden test set.

{\small
\bibliographystyle{ieeenat_fullname}
\bibliography{refs}
}

\end{document}